\documentclass[]{ceurart}
\begin{document}

\copyrightyear{2023}
\copyrightclause{Copyright for this paper by its authors.
  Use permitted under Creative Commons License Attribution 4.0
  International (CC BY 4.0).}

\conference{CLEF 2023: Conference and Labs of the Evaluation Forum, September 18–21, 2023, Thessaloniki, Greece}

\title{ARC-NLP at PAN 2023: Transition-Focused Natural Language Inference for Writing Style Detection}

\author[1]{Izzet Emre Kucukkaya}[
orcid=0009-0006-2877-8713,
email=ekucukkaya@aselsan.com.tr,
]
\address[1]{Aselsan Research Center, 06378, Ankara, Turkey.}

\author[1]{Umitcan Sahin}[
orcid=0000-0001-9594-3148,
email=ucsahin@aselsan.com.tr,
]

\author[1]{Cagri Toraman}[
orcid=0000-0001-6976-3258,
email=ctoraman@aselsan.com.tr,
]

\begin{abstract}
  The task of multi-author writing style detection aims at finding any positions of writing style change in a given text document. We formulate the task as a natural language inference problem where two consecutive paragraphs are paired. Our approach focuses on transitions between paragraphs while truncating input tokens for the task. As backbone models, we employ different Transformer-based encoders with warmup phase during training. We submit the model version that outperforms baselines and other proposed model versions in our experiments. For the easy and medium setups, we submit transition-focused natural language inference based on DeBERTa with warmup training, and the same model without transition for the hard setup.
\end{abstract}


\maketitle

\section{Introduction}
Multi-author writing style analysis is an important task due to recent advancements in AI-assisted conversational systems and chatbots. Moreover, it can be used for author verification and plagiarism detection.
In PAN 2023 \cite{bevendorff:2023}, participants are asked to find all positions of writing style change on the paragraph-level for a given text \cite{zangerle:2023}. In contrast to previous years of PAN, in this year, the style change happens in the same topic, which makes it challenging to detect writing style change by topic.

In this study, as Aselsan Research Center - Natural Language Processing team (ARC-NLP), we propose transition-focused natural language inference (NLI) for multi-author writing style detection. We formulate the task as a natural language inference problem where two consecutive paragraphs are paired and learning objective is to label whether they are written by the same author. We choose this approach since we already demonstrated that the NLI method performs well in similar tasks \cite{Toraman22}. As backbone models, we employ different Transformer-based language models with warmup phase during training (e.g. BERT-like encoders \cite{Devlin:2019}). Since the models have a limited length of input sequence (e.g. 512 tokens), we need to truncate the input paragraphs. For truncation, we focus on transitions between paragraphs, since transitions provide logical connections between paragraphs in documents. We are requested to send our models and algorithms as a docker image to the TIRA \cite{froebe:2023} system where the test set evaluation is made. 

\section{Related Work}
In PAN 2018 \citep{Kestemont:2018}, the task is basically binary classification. \citet{Zlatkova:2018} develop an ensemble approach of the models including SVM, Random Forest, LightGBM etc. \citet{Hosseinia:2018} use parallel attention networks to focus on the hierarchical structure of the language.

In PAN 2019 \cite{Zangerle:2019}, the task is to detect the number of authors in a given document. \citet{Nath:2019} use two clustering algorithms based on the threshold and window merge. In addition, \citet{Zuo:2019} use K-means and hierarchical clustering algorithms.

In PAN 2020 \cite{Zangerle:2020}, the task is to detect style changes between two consecutive paragraphs. \citet{Castro:2020} use a paragraph representation based on character, lexical, and syntactic features in a clustering algorithm. \citet{Iyer:2020} use a pre-trained BERT model, and train a random forest classifier of the embedding representation generated from the BERT model.

In PAN 2021 \cite{Zangerle:2021}, the task is to determine the number of authors, and locate specific author changes. \citet{Strom:2021} apply a stacking ensemble on text embeddings. \citet{Deibel:2021} use an LSTM-based algorithm. 

For the task of the last year, in PAN 2022 \citep{Zangerle:2022}, the winning solution \citet{Lin:2022} employ an ensemble of three Transformer-based language models using majority voting to obtain the final prediction. Furthermore, \citet{Jiang:2022} use base and large versions of the ELECTRA model, and report highly challenging scores. 

\begin{figure}[t]
    \centering
    \includegraphics[width=0.75\linewidth]{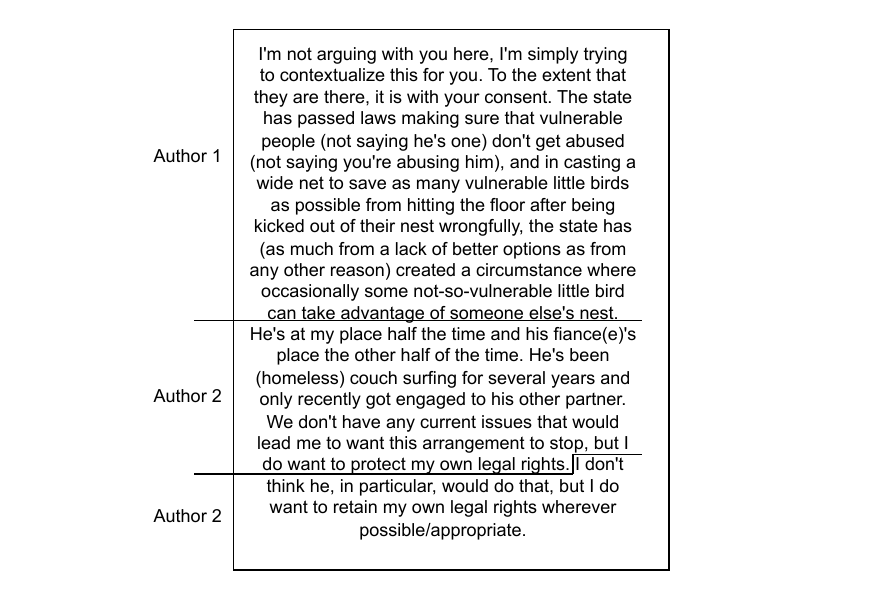}
    \caption{An example of writing style change. The labels are [1, 0] for this instance, representing writing style change in two consecutive paragraphs.}
    \label{fig:examples}
\end{figure}

\section{Task}
Participants are asked to solve the the intrinsic style change detection task. For a given text, we find all positions of writing style change on the paragraph-level. For example, the document in Figure \ref{fig:examples} is written by two authors, and there is a style change between first and second paragraph. The label of this transition is specified as 1. Furthermore, there is no style change between second and third paragraph where the label is 0. This example is chosen from the Easy split of the dataset. There are three difficulty levels:

\begin{itemize}
    \item \textbf{Easy}: The paragraphs of a document consist of various number of topics.
    \item \textbf{Medium}: The topical variety is small. The need of the style detection instead of topic detection increases. 
    \item \textbf{Hard}: All paragraphs in a document are on the same topic. 
\end{itemize}

All the documents in this task are in English, and contain different numbers of style changes and authors. Furthermore, writing style change only occurs in paragraph level. There is no need to investigate sentences separately.

\begin{table}[t]
\caption{Number of instances on the train and validation sets.}
\label{tab:dataset-stats}
\centering
\begin{tabular}{l c c}
\textbf{Difficulty} & \textbf{Train} & \textbf{Validation}\\
\hline
Easy & 4200 & 900\\
Medium & 4200 & 900\\
Hard & 4200 & 900\\

\end{tabular}
\end{table}

\section{Dataset}
\label{sec:dataset}
In this task, there are three different difficulty levels with their own train-validation-test sets. The numbers of documents on the train and validation splits are the same in all of the three difficulty levels, which are provided in Table \ref{tab:dataset-stats}.

The total number of the 0 and 1 labels in documents are reported in Table \ref{tab:dataset-v2}. These numbers indicate the number of samples in the natural language inference task derived from the actual task as mentioned in Section \ref{sec:method}.

\begin{table}[t]
\caption{Number of pairs for each label. The 0 and 1 labels indicate that if the style change occurs.}
\label{tab:dataset-v2}
\centering
\begin{tabular}{l c | c c | c}
 \multirow{2}{*}{\textbf{Difficulty}} & \multicolumn{2}{c}{\textbf{Train}} & \multicolumn{2}{c}{\textbf{Validation}}\\
 & \textbf{0} & \textbf{1} & \textbf{0} & \textbf{1} \\
\hline
Easy &   1557  & 11347 & 377  & 2451 \\
Medium & 15001  & 13215 & 4013 & 3029 \\
Hard & 10092  & 9021 & 2159  & 1953  \\
\end{tabular}
\end{table}

\section{Proposed Method: Transition-Focused Natural Language Inference}
\label{sec:method}
\begin{figure}[h]
    \centering
    \includegraphics[trim={0 3.5cm 0 1cm},clip]{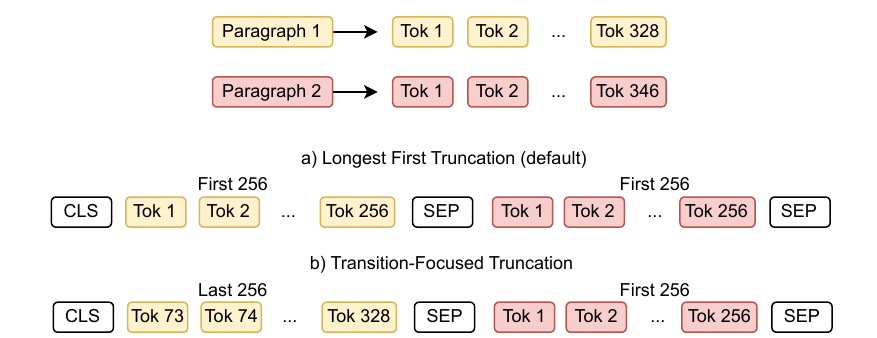}
    \caption{An illustration of the proposed approach that uses natural language inference with transition-focused truncation.}
    \label{fig:truncation}
\end{figure}

\subsection{Main Approach}
In this work, we formulate the task as natural language inference (NLI). To do so, we employ a Transformer-based language model that is based on the encoder structure. We prepare input by concatenating consecutive paragraphs and using the SEP token between them. We then place a binary classification layer of the CLS embedding, based on whether if the style change occurs (1) or not (0) between these to paragraphs.  

Since the models have a limited length of input sequence (i.e. 512 tokens for BERT, RoBERTa, and DeBERTa; and 1024 tokens for BigBird), we need to truncate the input paragraphs before training NLI. For truncation, we focus on transitions between paragraphs (we refer it to as \textbf{Transition-Focused Truncation)}, since transitions provide logical connections between paragraphs in documents. We also provide results for default truncation that focuses on the beginning of text (we refer it to as \textbf{Longest First Truncation)}. The proposed NLI model and truncation approaches are illustrated in Figure \ref{fig:truncation}. When input sequence length is 512, the last 256 tokens of the first paragraph and the first 256 tokens of the second paragraph are combined in transition-focused truncation as in Figure \ref{fig:truncation}a, while first 256 tokens from both paragraphs are truncated as in Figure \ref{fig:truncation}b.

\subsection{Backbone Models}
As the text encoder, we employ several Transformer-based language models in our preliminary experiments. Here, we report the highest performing four models.

\paragraph{BERT \cite{Devlin:2019}}
Bidirectional encoder representation for transformers, BERT, is an encoder architecture that utilizes an attention mechanism. It was pretrained on masked language modelling and next sentence prediction tasks.

\paragraph{RoBERTa \cite{Liu:2019}}
A robust optimized BERT pre-training approach, RoBERTa, has the same architecture as BERT. However, the pretraining task of the next sentence prediction is removed. Furthermore, it has dynamically changing masking pattern applied to the training data with larger training batches.

\paragraph{DeBERTa v3 \cite{He:2021}}
Decoding-enhanced BERT with Disentangled Attention, DeBERTa, has two additional techniques compared to the BERT, distangled attention and enhanced mask decoder. Due to the new adjustments, they state that it outperforms BERT and the other state-of-art models in many tasks.

\paragraph{BigBird-RoBERTa \cite{Zaheer:2020}}
 BigBird has a sparse attention mechanism that reduces this quadratic dependency to linear which enables it to handle sequences of length up to 8 times of what was previously possible using similar hardware. Since paragraphs can be too long in this task, we employ this model to cover more number of tokens in input.

\subsection{Warmup}
In preliminary experiments, we realize that our models converge different minima in the training of this task. In order to overcome this issue, we use the warmup with the warmup ration is 0.1. In warmup steps, the model trains with a very low learning rate and tries to find the global minima of the loss function, and hinders the inaccurate convergence.

\section{Experiments}

\subsection{Experimental Setup}
The input length in BigBird is 1024 tokens (512 for each paragraph), while 512 tokens for other models (256 for each paragraph). We set the following hyperparameters. Learning rate is 5e-5, number of epochs is 5, and batch size is 4. We used 3 NVIDIA GeForce RTX 2080 GPUs in training. The pre-trained models and the trainer framework are obtained from the HuggingFace library \cite{wolf:2020}. 

For evaluating model performances, we calculate Macro F1 Scores using the official evaluation  script\footnote{\href{https://github.com/pan-webis-de/pan-code/tree/master/clef23/multi-author-analysis}{https://github.com/pan-webis-de/pan-code/tree/master/clef23/multi-author-analysis}}.

\subsection{Baselines}
We implement two baseline methods to compare with our approach.

\paragraph{Random}
The output label array is generated randomly by sampling from 0 and 1, uniformly.

\paragraph{TF-IDF}
We use TF-IDF term weighting \cite{Salton:1984} to extract features using the English stopwords of NLTK library \cite{bird:2004}. Additional features such as number of question marks, periods, apostrophes, parenthesis, and words are concatenated to the feature vector. Finally, we concatenate the TF-IDF feature vectors of consecutive two paragraphs, and train Support Vector Classifier (SVC) \cite{Cortes:1995} for classification.

\begin{table}[t]
\caption{Macro F1 Scores on the validation set. We submitted the highest performing models on the validation set given in bold. The test (leaderboard) results are given at the end of the table.}
\label{tab:results}
\centering
\begin{tabular}{l c c c}
\textbf{Model} & \textbf{Easy} & \textbf{Medium} & \textbf{Hard} \\
\hline
random & 0.401 & 0.504 & 0.505 \\
tfidf & 0.551 & 0.615 & 0.531\\
\hline
bert-base & 0.980 & 0.807 & 0.753 \\
roberta-base & 0.790 & 0.363 & 0.344 \\
deberta-v3 & 0.966 & 0.591 & 0.344 \\
bigbird-roberta-base & 0.965 & 0.400 & 0.344 \\
\hline
bert-base-warmup & 0.982 & 0.786 & 0.755\\
roberta-base-warmup & 0.662 & 0.766 & 0.643\\
deberta-v3-warmup & 0.984 & 0.798 & \textbf{0.770}\\
bigbird-roberta-base-warmup & 0.962 & 0.805 & 0.742\\
\hline
bert-base-transition & 0.918 & 0.766 & 0.344 \\
roberta-base-transition & 0.976 & 0.363 & 0.344 \\
deberta-v3-transition & 0.464 & 0.778 & 0.344 \\
bigbird-roberta-base-transition & 0.969 & 0.796 & 0.724\\
\hline
bert-base-warmup-transition & 0.958 & 0.772 & 0.660\\
roberta-base-warmup-transition & 0.981 & 0.804 & 0.344 \\
deberta-v3-warmup-transition & \textbf{0.987} & \textbf{0.812} & 0.746 \\
bigbird-roberta-base-warmup-transition & 0.958 & 0.804 & 0.759\\
\hline
\emph{Submitted models on the test set} & \emph{0.982} & \emph{0.810} & \emph{0.772}\\
\end{tabular}
\end{table}

\subsection{Experimental Results}

We report the model performances on the validation splits (see Section \ref{sec:dataset}) in Table \ref{tab:results}. We divide the table into six parts. At the top, we provide the baseline scores. The second part consists of our proposed approach with four backbone models, described in Section \ref{sec:method}. We use the same models with warmup during training in the third part. So far, we do not employ transition-focused truncation. Next, we provide the results of transition-focused truncation and lastly with warmup as well. In the last part, we provide the performance scores of the submitted models on the test set (leaderboard). We submitted the highest performing models on the validation set (given as bold). We have the following observations. 

\begin{itemize}
    \item Baseline models perform poor as expected. TF-IDF is based on bag-of-words model, which can show that writing style can not be detected by syntactical writing features.
    \item DeBERTa is the highest performing backbone model for NLI in all setups.
    \item Using warmup during training can increase the performance in some cases, specifically for the medium and hard setups.
    \item Transition-focused truncation method improves the results in some cases. More importantly, we obtain the highest scores on the validation set for the easy and medium setups when we employ transition-focused DeBERTa with warmup. For the hard setup, the same model with default truncation performs highest. We submitted the highest performing methods to the shared task.
\end{itemize}

\section{Conclusion}
In this paper, we propose transition-focused natural language inference (NLI) for multi-author writing style detection. We truncate the input paragraphs by focusing on transitions between paragraphs, since transitions provide logical connections between paragraphs in documents. Transition-focused NLI performs highest in easy and medium setups. Moreover, we obtain the highest performances when backbone model is DeBERTa in all setups. We submitted the highest performing models on the validation set. Our models are placed in the second place for all subtasks (easy, medium, and hard) in the leaderboard.

As a future work, there can be some improvements to overcome the class imbalance problem. Furthermore, other large language models can be employed to encode the embedding vectors.

\bibliography{pan}

\appendix

\end{document}